\documentclass[sigconf, 
                screen, 
                ]{acmart}

\AtBeginDocument{%
  \providecommand\BibTeX{{%
    \normalfont B\kern-0.5em{\scshape i\kern-0.25em b}\kern-0.8em\TeX}}}

\usepackage{multirow}
\usepackage{xcolor}
\usepackage{subfigure}
\usepackage{algorithm}
\usepackage{algorithmic}
\usepackage{balance}

\def\calc{\mathcal{C}}
\def\calt{\mathcal{T}}

\def\calg{\mathcal{G}}
\def\allclasses{\caly_\calg}

\DeclareMathOperator*{\argmax}{argmax} 
\DeclareMathOperator*{\argmin}{argmin}

\def\calx{\mathcal{X}}
\def\caly{\mathcal{Y}}

\def\fineclasses{\caly_f}
\def\coarseclasses{\caly_c}

\def\assign{\mathit{assign}}
\def\error{\mathit{error}}

\newcommand{\trainedpredictions}[1]{\hat{Y}_{\hat{\theta}}^{#1}}

\def\trainedtpredictions{\trainedpredictions{\calt}}

\def\thetahat{\hat{\theta}}
\def\fthetahatmodel{f_{\hat{\theta}}}

\copyrightyear{2024}
\acmYear{2024}
\setcopyright{acmlicensed}
\acmConference[CIKM '24]{Proceedings of the 33rd ACM International Conference on Information and Knowledge Management}{October 21--25, 2024}{Boise, ID, USA}
\acmBooktitle{Proceedings of the 33rd ACM International Conference on Information and Knowledge Management (CIKM '24), October 21--25, 2024, Boise, ID, USA}
\acmDOI{10.1145/3627673.3679918}
\acmISBN{979-8-4007-0436-9/24/10}
\begin{document}
\title{Error Detection and Constraint Recovery in Hierarchical Multi-Label Classification without Prior Knowledge}

\author{Joshua Shay Kricheli}
\orcid{0000-0001-8398-6378}
\affiliation{%
  \institution{Arizona State University}
  \city{Tempe}
  \state{Arizona}
  \country{USA}
  }
\email{jkrichel@asu.edu}

\author{Khoa Vo}
\orcid{0009-0007-2894-3616}
\affiliation{%
  \institution{Arizona State University}
  \city{Tempe}
  \state{Arizona}
  \country{USA}
  }
\email{ngocbach@asu.edu}

\author{Aniruddha Datta}
\orcid{0009-0002-1619-4586}
\affiliation{%
  \institution{Arizona State University}
  \city{Tempe}
  \state{Arizona}
  \country{USA}
}
\email{adatta14@asu.edu}

\author{Spencer Ozgur}
\orcid{0009-0007-8151-2233}
\affiliation{%
  \institution{Arizona State University}
  \city{Tempe}
  \state{Arizona}
  \country{USA}
  }
\email{sozgur@asu.edu}

\author{Paulo Shakarian}
\orcid{0000-0002-3159-4660}
\affiliation{%
  \institution{Arizona State University}
  \city{Tempe}
  \state{Arizona}
  \country{USA}
  }
\email{pshak02@asu.edu}

\renewcommand{\shortauthors}{Joshua Shay Kricheli, Khoa Vo, Aniruddha Datta, Spencer Ozgur, \& Paulo Shakarian}

\begin{abstract}
Recent advances in Hierarchical Multi-label Classification (HMC), particularly neurosymbolic-based approaches, have demonstrated improved consistency and accuracy by enforcing constraints on a neural model during training. However, such work assumes the existence of such constraints a-priori. In this paper, we relax this strong assumption and present an approach based on Error Detection Rules (EDR) that allow for learning explainable rules about the failure modes of machine learning models. We show that these rules are not only effective in detecting when a machine learning classifier has made an error but also can be leveraged as constraints for HMC, thereby allowing the recovery of explainable constraints even if they are not provided. We show that our approach is effective in detecting machine learning errors and recovering constraints, is noise tolerant, and can function as a source of knowledge for neurosymbolic models on multiple datasets, including a newly introduced military vehicle recognition dataset.
\end{abstract}

\ccsdesc[500]{Computing methodologies~Knowledge representation and reasoning}
\ccsdesc[500]{Computing methodologies~Rule learning}
\ccsdesc[500]{Computing methodologies~Logical and relational learning}

\keywords{Metacognitive AI, Learning with Constraints, Rule Learning, NeuroSymbolic AI, Hierarchical Multi-label Classification}

\maketitle

\section{Introduction}
Hierarchical Multi-label Classification (HMC) extends the idea of multi-label classification to impose a hierarchy among labels~\cite{giunchiglia2020coheren, pmlr-v80-wehrmann18a, 9894725, decisionVens} and has been applied to a variety of applications~\cite{zhongchenma2022multilabel, ltn22, liu2023recent}. A separate line of machine learning research deals with the training of a secondary model to identify failures in the first~\cite{daftry_introspective_2016,ramanagopal_failing_2018,harrison22,NEURIPS2023_fe318a2b} - which has been referred to as ``\textit{metacognition}''~\cite{Maes1988MetaLevelAA,JOHNSON2022105743} and typically the secondary model is a black-box model. Recent research trends (specifically neurosymbolic AI~\cite{nsaiBook23, thirdWave2020}) have led to the expression of such hierarchical relationships as constraints on the learning process~\cite{ltn22,xu2018semantic,giunchiglia2020coheren} or leveraging constraints to make corrections to the machine learning model~\cite{cornelio2022learning,Huang_Dai_Jiang_Zhou_2023}.  A key assumption in these approaches is that the constraints are known \textit{a-priori}. Meanwhile, a metacognitive approach known as Error Detection Rules (EDR)  allows for the learning of rules to predict failures~\cite{xi2023rulebased} based on conditions derived from domain knowledge and/or complementary models for the same task trained on the same data.  The key intuition of this paper is that we can modify the ideas of \cite{xi2023rulebased} to predict errors and recover constraints of an HMC model without a-prior knowledge of the constraints. Our contributions are as follows: (1) We extend the EDR framework of \cite{xi2023rulebased} to address the HMC problem without prior knowledge of the hierarchy by both extending the language of that work and presenting a new \textit{Focused-EDR} which addresses an objective function mismatch of \cite{xi2023rulebased} by leveraging approximate optimization of the ratio of two submodular functions, (2) We demonstrate how our new approach, provides significant improvement in the detection of errors when compared to a black-box baseline neural error detector and the detection algorithm of \cite{xi2023rulebased} on three different HMC datasets, (3) We show our approach can recover constraints and that both the F1-score of constraints recovered as well as error F1 degrades gracefully with noise - with noise injected in a manner to remove certain classes from consideration, (4) We show the recovered constraints can then be used as a source for in neurosymbolic model learning (i.e., Logic Tensor Networks (LTN)~\cite{ltn22}) to provide improved model performance and reduce the number of inconsistencies and (5) We introduce (and release \footnote{\url{https://github.com/lab-v2/PyEDCR}}) an open-source HMC dataset of $\sim$10K images of military vehicles.
The rest of the paper is organized as follows. We provide an overview of related work in Section~\ref{sec:related_work}, introduce our Focused EDR approach in Section~\ref{sec:approach}, followed by our experimental results and discussion in Section~\ref{sec:exp_results}.
\section{Related Work}
\label{sec:related_work}

\begin{table*}
\caption{Example Error Detection rules for a label $y$ and violating sets $V_y$ for several fine and coarse-grain classes ($\caly_f$ and $\caly_c$ respectively) of military vehicles. A violating set $V_y$ is a set of labels which are inconsistent with assigning the class $y$.} 
 \begin{tabular}{cccc}
    \toprule
    \multicolumn{1}{c}{Granularity $g$} & 
    \multicolumn{1}{c}{Label $y$}&
    \multicolumn{1}{c}{Error Detecting Rule}&
    \multicolumn{1}{c}{Violating Set $V_y$}\\
    \midrule
    \multirow{2}{*}{Coarse-Grain}&  
    $\textit{Tank}$& 
    $\error_{\textit{Tank}}(X) \leftarrow \assign_{\textit{Tank}}(X) \wedge \bigvee_{y' \in V_y}\assign_{y'}(X)$ &
    $\fineclasses \setminus \{ \textit{T-14},\textit{T-62},\textit{T-64},\textit{T-72},\textit{T-80},\textit{T-90}\}$
    \\ 
    & $\textit{SPA}$&
    $\error_{\textit{SPA}}(X) \leftarrow \assign_{\textit{SPA}}(X) \wedge \bigvee_{y' \in V_{y}}\assign_{y'}(X)$ &
    $\fineclasses \setminus \{\textit{2S19-MSTA},\textit{BM-30},\textit{D-30},\textit{Tornado},\textit{TOS-1}\}$
    \\ 
    \multirow{2}{*}{Fine-Grain}&  
    $\textit{T-14}$& $\error_{\textit{T-14}}(X) \leftarrow \assign_{\textit{T-14}}(X) \wedge \bigvee_{y' \in V_{y}}\assign_{l'}(X)$ &
    $\coarseclasses \setminus \{\textit{Tank}\}$
    \\ 
    &  
    $\textit{2S19-MSTA}$& $\error_{\textit{2S19-MSTA}}(X) \leftarrow \assign_{\textit{2S19-MSTA}}(X) \wedge \bigvee_{y' \in V_{y}}\assign_{y'}(X)$ &
    $\coarseclasses \setminus \{\textit{SPA}\}$\\ 
    \bottomrule
  \end{tabular}
\label{tbl:example_error_detection_rules}
\end{table*}
Related work includes HMC~\cite{9894725,decisionVens,noor2022capsule}, neurosymbolic AI~\cite{xu2018semantic,giunchiglia2020coheren,ltn22}, metacognition ~\cite{Maes1988MetaLevelAA,JOHNSON2022105743,werner24, dai2019bridging,huang2023enabling,cornelio2022learning} and conformal prediction~\cite{angelopoulos2022gentle, NEURIPS2023_fe318a2b, angelopoulos2022uncertainty}. Error Detection Rules (EDR) was introduced in~\cite{xi2023rulebased} and is a metacognitive approach for detecting errors in the result of a trained machine learning model $\fthetahatmodel$ that assigns a label $y$ for some sample $x$. In that work, the logical atom $\assign_y(X)$ means that the model assigned label $y$ to a sample denoted by the variable symbol $X$.  Additionally, the method takes a set of boolean conditions $\calc$ as input, which are associated with each sample (e.g., $cond(X) \equiv True$, iff condition $cond$ is true for sample $x$). In practice, these conditions can come from domain knowledge or from a complementary model for the same task (as done in \cite{xi2023rulebased}).  The key intuition is for each class $y$ to identify a subset of conditions $DC_y \subseteq C$ that are indicative of an error made by the model.  This results in an \textit{error detection rule}:
\begin{equation}
\label{rule:xi-det}
\error_y(X) \leftarrow \assign_y(X) \wedge \bigvee_{cond \in DC_y}cond(X).
\end{equation}
Intuitively, this reads that if one of the conditions in set $DC_y$ are observed for sample $X$ and the model assigns class $y$ to it, then the model has made an error predicting $x$ to be of class $y$. The authors learn the rules through a constrained submodular optimization technique that optimizes a target class precision.

\section{Approach}
\label{sec:approach}
We now describe some technical preliminaries for our problem and two key extensions we make to the EDR framework of~\cite{xi2023rulebased}.  Specifically, we extend the EDR method to capture the HMC case and we address the objective function mismatch in the detection algorithm. In our setup, we assume that the hierarchical data has $G$ levels of granularities $\calg \coloneqq \{g_i\}_{i=1}^G$, each with a corresponding label set $\caly_g$. In this work we focus in the case of $G=2$, and denote $\calg \coloneqq \{fine, coarse\}$. The framework can extend for $G > 2$, but we leave evaluation of the approach beyond two levels to future work. Let $\calx, \caly$ be all the possible examples and predicted labels, respectively. We thus have $\caly = \bigtimes_{g\in\calg}\caly_g$ as all the possible predicted labels, and define $\allclasses \coloneqq \bigcup_{g\in\calg}\caly_g$. Define a labeled training set $\calt \coloneqq \{(x^{\calt}_i,gt(x^{\calt}_i))\}_{i=1}^{N_{\calt}} \subset \calx \times \caly$ composed of $N_{\calt}$ images and corresponding ground-truth labels. We assume the existence of a well-trained model $\fthetahatmodel$ returning one class per level of granularity for a given sample. We then define an \textit{Error Prediction Problem} per class $y\in \caly_\calg$, which is predicting where $\fthetahatmodel$ predicted $y$ incorrectly. We envision such a predictor to be trained on the same data as $\fthetahatmodel$ and provide a per-example binary output for $error_y(X)$ (the rule head of~\eqref{rule:xi-det}), which we refer to as the error class of $y$, and denote it $e_y$. Earlier work such as~\cite{daftry_introspective_2016} relied on a black box model for error predictions, while this paper along with the recent work of~\cite{xi2023rulebased} utilizes one set of rules per class, as in Expression~\eqref{rule:xi-det}.

\noindent \textbf{EDR for HMC.} The key difference in employing EDR for HMC problems is that we are detecting errors in $G > 1$ categories of classes which have hierarchy constraints among them. Hence, we adjust the rules of the form given in Expression~\eqref{rule:xi-det} for an HMC problem as well as support conditions from multiple models.  First, we extend the predicate $\assign$ where we not only have a subscript denoting the class (which can come from multiple granularities) but also a superscript denoting the model, so $\assign_y^{f}(X)$ means that sample $x$ was assigned label $y$ by model $f$. Note that the first predicate in the body of the rule (outside the disjunction, as per Expression~\eqref{rule:xi-det}) will always come from the base model.  While~\cite{xi2023rulebased} primarily relied on domain knowledge for the condition sets, here we use the class predictions from a different label set of the hierarchy. As such, when considering detection rules for any class $y$, as in~\eqref{rule:xi-det}, we will consider conditions from the model predictions defined using granularities other than that of $y$. Thus for each class $y$ at granularity $g$, we define a specific conditions set $\calc_g$ (unlike in~\cite{xi2023rulebased} where conditions are picked from the same set of conditions as there were no levels of granularity in that paper). The per-granularity condition set will include $\assign$ conditions from the `\textit{main}' model $\fthetahatmodel$ across all other granularities, which are $\calc^{\fthetahatmodel}_g \coloneqq \bigcup_{g'\in\calg\setminus \{ g\}}\bigcup_{y\in \caly_{g'}} \{\assign^{\fthetahatmodel}_y\}$. Since each such $\calc^{\fthetahatmodel}_g$ will not include conditions on labels from $g$, we also add conditions from a `\textit{secondary}' model $h_{\thetahat_h}$, with a possible different architecture $h$ and learned parameters $\thetahat_h$, for all granularities, i.e. $\calc^{h_{\thetahat_h}} \coloneqq \bigcup_{g\in \calg}\bigcup_{y\in \caly_{g'}} \{\assign^{h_{\thetahat_h}}_y\}$. Moreover, similar to what the authors did in~\cite{xi2023rulebased}, we define binary conditions for each label of each granularity which utilize trained binary models $\{b_{\thetahat_y}\}_{y\in B}$, given a subset of labels $B \subseteq \allclasses$, yielding the binary conditions set $\calc_B \coloneqq \bigcup_{y\in B} \{\assign^{b_y}\}$. Finally, we set the condition set for each granularity to $\calc_g \coloneqq \calc^{\fthetahatmodel}_g \cup \calc^{h_{\thetahat_h}} \cup \calc_B$. We note that in solving the error detection problem for a given label $y$, we find a set $DC_y$ to create a rule of the form seen in Expression~\eqref{rule:xi-det}. Notice how after solving for all the sets $\{DC_y\}_{y\in{\caly_\calg}}$ for an HMC problem, for any granularity $g$ and label $y\in \caly_g$, the conditions in $DC_y \cap \calc^{\fthetahatmodel}_g$ are specifying hierarchy constraints, i.e. rules of the form ``if label $y$ is assigned by the model at granularity $g$ and another label $y'$ at a different granularity $g' \neq g$, then there is an error in assigning $y$.'' These rules enforce hierarchical consistency, as defined by the ground truth labels. Examples of such rules are given in Table~\ref{tbl:example_error_detection_rules}.

\noindent\textbf{Optimization Approach.} As mentioned earlier, in the work of~\cite{xi2023rulebased}, the algorithm that learns the set $DC_y$ is done so in a way to optimize the precision of $y$ as opposed to optimizing a metric to detect errors (e.g., as done in~\cite{daftry_introspective_2016}). Our key intuition for our optimization approach is that for each label $y$, we wish to maximize the F1-score of the error class $e_y$ (as defined in the \textit{Error Prediction Problem}) as opposed to precision. In this way, we will not only improve error detection, but also recover constraint violations that lead to errors. Our strategy to maximize the error F1 is to identify an equivalent quantity and prove this quantity is the ratio of two submodular functions. This, in-turn allows us to leverage the approximation routine of~\cite{pmlr-v48-baib16} which we later show provides good results in practice. Following the convention of~\cite{xi2023rulebased} we define $POS^{\calt}_{DC_y}$ as the number of examples in $\calt$ that satisfy any of the conditions in $DC_y \subseteq \calc$, are classified as $y \in \caly_{\calg}$ by $\fthetahatmodel$ and are also False Positives in the prediction vector of $\fthetahatmodel$ on $\calt$, which we will denote $\trainedtpredictions$. We define $BOD^{\calt}_{DC_y}$ as the number of examples in $\calt$ that satisfy any of the conditions in $DC_y$ and were predicted as $y$ in $\trainedtpredictions$, and $FP^{\calt}_y$ is the number of False Positives of class $y$ in $\trainedtpredictions$. For each class $y$, we then seek to find an approximate solution to the following optimization problem:
\begin{equation}
    \begin{aligned}
    \label{eqn:ratio_maximization}
    DC^*_y \in & \argmax_{\emptyset \subset DC_y \subseteq \calc}\frac{POS^{\calt}_{DC_y}}{BOD^{\calt}_{DC_y}+ FP^{\calt}_y}.
    \end{aligned}
    \end{equation}
We note $POS^{\calt}_{DC_y}$ is equivalent to the number of True Positives of $e_y$ (the error class of $y$) in the training set $\calt$, $BOD^{\calt}_{DC_y}$ is the sum of the True and False Positives of $e_y$ in $\calt$ and $FP^{\calt}_y$ is the sum of the True Positives and False Negatives of $e_y$ in $\calt$. Using these, the proof that the quantity in~\eqref{eqn:ratio_maximization} is equal to the F1-score of $e_y$ in $\calt$ is straightforward. We provide the formal statement below.

\begin{theorem}
Given a training set $\calt$, a solution $DC^*_y$ for the maximization problem in Equation~\eqref{eqn:ratio_maximization} of a class $y \in \allclasses$ also maximizes the error \textsf{F1}-score for class $y$ with respect to the prediction vector $\trainedtpredictions$.
\end{theorem}

\noindent We note that the quantities of $POS^{\calt}_{DC_y}$ and $BOD^{\calt}_{DC_y}$ are both submodular by Claim 2 of Theorem 4 of~\cite{xi2023rulebased} while $FP^{\calt}_y$ is a constant for any choice of $DC_y$.  Hence, the quantity in Expression~\eqref{eqn:ratio_maximization} is the ratio of two submoular functions, which allows us to leverage a variant of Algorithm 3 from~\cite{pmlr-v48-baib16}. Notice how this algorithm solves for minimizing a ratio of submodular functions $f/g$, when the authors of~\cite{pmlr-v48-baib16} explain in the introduction how the same solution solves the problem of maximizing for $g/f$. Thus, we maximize for the ratio in Expression~\eqref{eqn:ratio_maximization} by minimizing for its inverse. The modified algorithm is described in Algorithm~\ref{alg:RatioDetRuleLearn} in this work. Referring to the runtime described in~\cite{pmlr-v48-baib16}, we note that in the worst case, this algorithm runs in a time complexity of $O(|\calc_g|^2)$, rendering it efficient and scalable as the number of conditions grow.

\begin{algorithm}
\caption{\textsf{RatioDetRuleLearn}
\label{alg:RatioDetRuleLearn}}
\begin{algorithmic}
\REQUIRE Class $y \in \caly_g$ of a granularity $g$ and its per-granularity condition set $\calc_g$
\ENSURE Non-empty subset of conditions $\emptyset \subset \hat{DC}_y \subseteq \calc_g$\\
\STATE{$DC^0_y \leftarrow \emptyset, \calc_y \leftarrow \calc_g, i \leftarrow 0$}
\WHILE{$\calc_y \neq \emptyset$}
    \STATE{$c_{best} \in \argmin_{c \in \calc_y} \frac{BOD^{\calt}_{DC^{i}_y \cup\{c\}} - BOD^{\calt}_{DC^{i}_y}}{POS^{\calt}_{DC^{i}_y \cup\{c\}} - POS^{\calt}_{DC^{i}_y}}$}
    \STATE{$DC^{i+1}_y \leftarrow DC^{i}_y \cup \{c_{best}\}$}
    \STATE{$\calc_y \leftarrow \Big\{ c \in \calc_y \ \Big| \  POS^{\calt}_{DC^{i+1}_y  \cup\{c\}} > POS^{\calt}_{DC^{i+1}_y  }\Big\}$}
    \STATE{$i \leftarrow i + 1$}
\ENDWHILE
\STATE{$\hat{i} \in \argmin_i \frac{BOD^{\calt}_{DC^{i}_y} + FP^\calt_y}{POS^{\calt}_{DC^{i}_y}}$}
\STATE{\textbf{return} $\hat{DC}_y \leftarrow DC^{\hat{i}}_y$}
\end{algorithmic}
\end{algorithm}

\begin{table*}
  \caption{Error Correction Results with LTN}
  \begin{tabular}{ccccccc}
    \toprule
    \multicolumn{1}{c}{Dataset} & \multicolumn{1}{c}{Method} &
    \multicolumn{1}{p{1.5cm}}{\centering Fine-Grain Acc.}& 
    \multicolumn{1}{p{1.5cm}}{\centering Fine-Grain f1}&
    \multicolumn{1}{p{1.5cm}}{\centering Coarse-Grain Acc.}& 
    \multicolumn{1}{p{1.5cm}}{\centering Coarse-Grain f1}&
    \multicolumn{1}{p{2cm}}{\centering Inconsistency}\\
    \midrule
    \multirow{2}{*}{Military Vehicles}& 
    \multicolumn{1}{c}{VIT b\_16} & 
    54.35\%& 
    48.40\%& 
    77.24\%&
    74.57\%&
    7.77\% (126/1621)\\
    &\multicolumn{1}{c}{f-EDR + LTN (ours)} &
    \textbf{62.43}\%& 
    \textbf{58.18}\%& 
    \textbf{82.17}\%&
    \textbf{79.95}\%&
    \textbf{5.37}\% (87/1621)\\
    
    \cline{1-7} \multirow{2}{*}{ImageNet50}& 
    \multicolumn{1}{c}{DINO V2 s\_14} &
    85.76\%& 
    85.59\%&
    93.52\%&
    92.65\%&
    1.19\% (25/2100) \\
    &\multicolumn{1}{c}{f-EDR + LTN (ours)} &
    \textbf{86.29}\%& 
    \textbf{86.12}\%&
    \textbf{93.57}\%&
    \textbf{92.69}\%&
    \textbf{1.05}\% (22/2100)\\
    
    \cline{1-7} \multirow{2}{*}{OpenImage36}& 
    \multicolumn{1}{c}{VIT b\_16} &
    57.68\%& 
    55.89\%&
    90.15\%& 
    88.82\%&
    3.02\% (362/12002)
    \\
    &\multicolumn{1}{c}{f-EDR + LTN (ours)} &
    \textbf{60.11}\%& 
    \textbf{58.88}\%& 
    \textbf{91.21}\%&
    \textbf{89.85}\%&
    \textbf{1.73}\% (208/12002)
    \\
    \bottomrule
  \end{tabular}
  \label{tbl:error_correction_results}
\end{table*}

\begin{figure*}[ht]
    \centering
    \begin{minipage}[t]{0.33\textwidth}
        \centering
        \includegraphics[width=\textwidth]{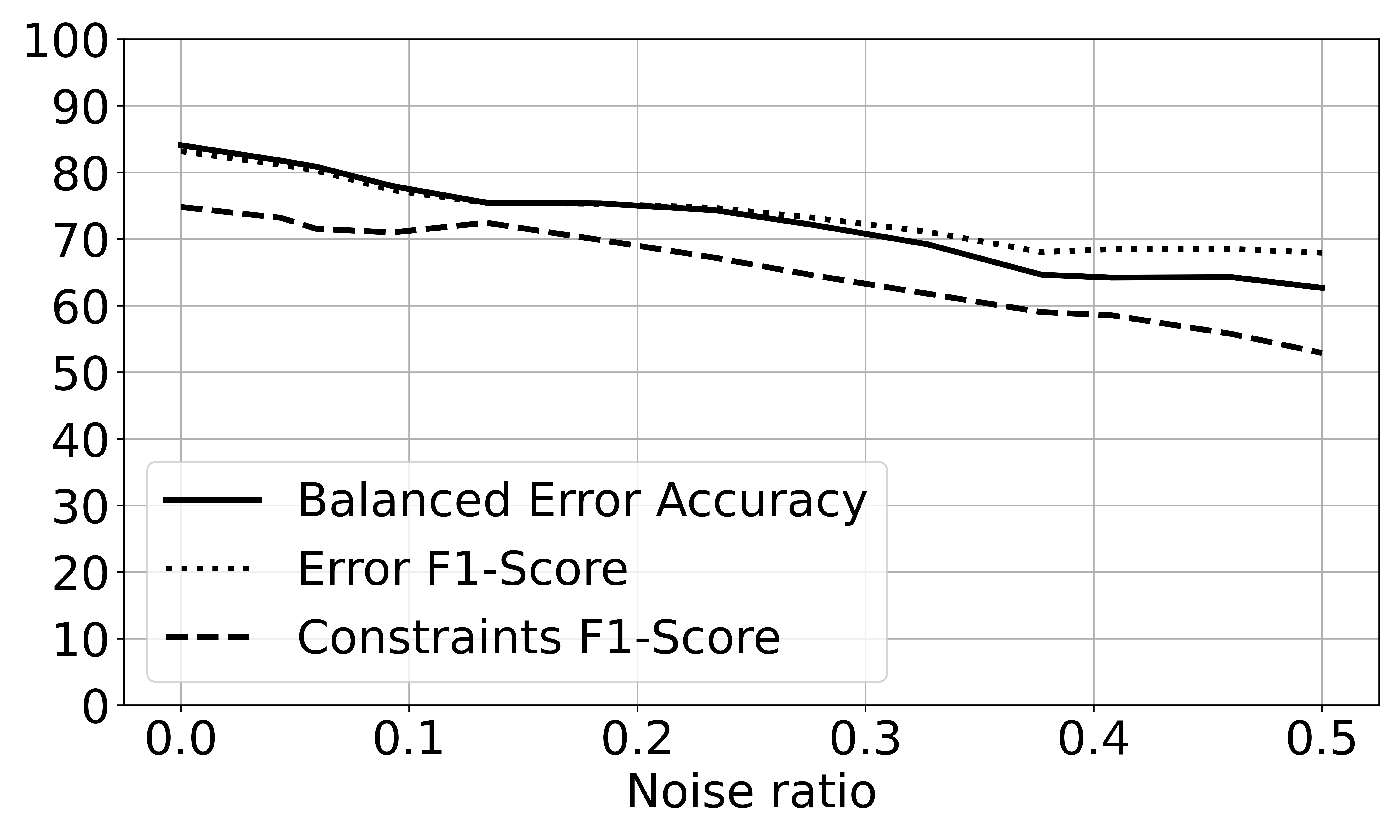}
        \caption*{(a)}
    \end{minipage}%
    \hfill%
    \begin{minipage}[t]{0.33\textwidth}
        \centering
        \includegraphics[width=\textwidth]{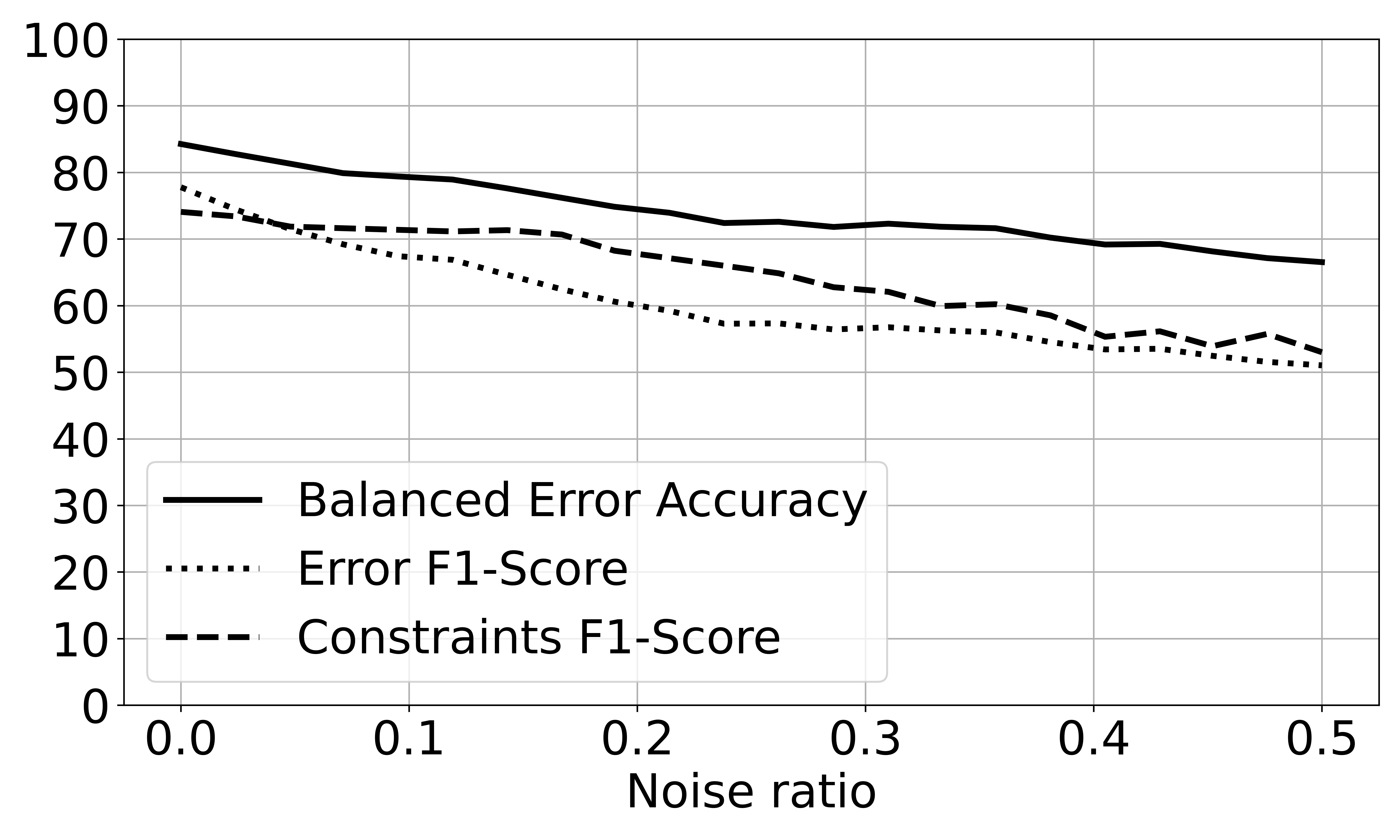}
        \caption*{(b)}
    \end{minipage}%
    \hfill%
    \begin{minipage}[t]{0.33\textwidth}
        \centering
        \includegraphics[width=\textwidth]{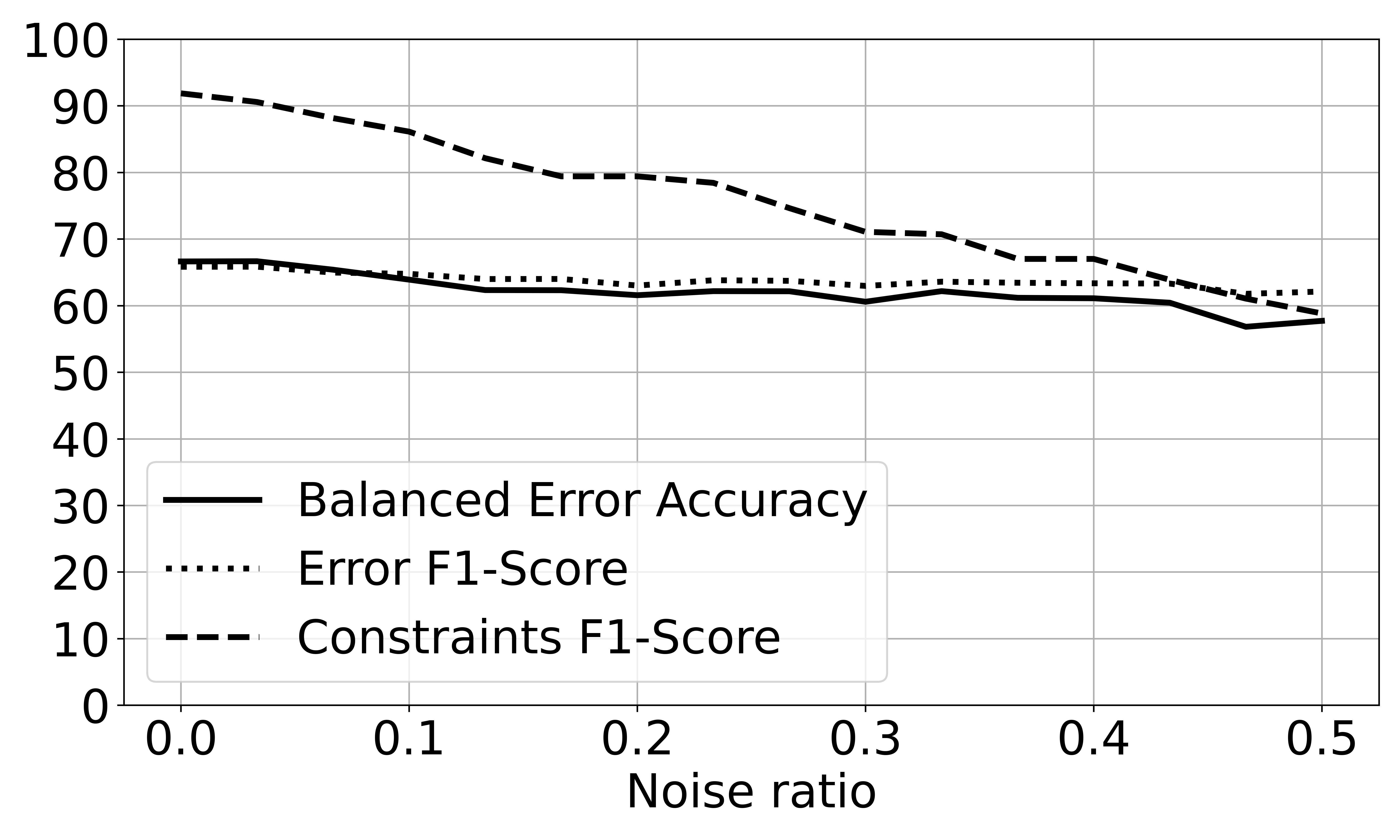}
        \caption*{(c)}
    \end{minipage}
    \caption{Balanced Error Accuracy, Error \textsf{F1}-score, and constraints \textsf{F1}-score results of the Focused-EDR method for varying noise ratios for the Military Vehicles (a) ImageNet50 (b) and OpenImage36 (c) datasets}
    \Description{Focused-EDR metrics for the experimented datasets.}
    \label{fig:OOD}
\end{figure*}
\begin{table}
  \caption{Error Detection Results for all the datasets}
  \begin{tabular}{cccc}
    \toprule
    \multicolumn{1}{c}{Dataset} & \multicolumn{1}{c}{Method} &
    \multicolumn{1}{p{1.6cm}}{\centering Balanced Error Acc.}& 
    \multicolumn{1}{p{1.6cm}}{\centering Error f1}\\
    \midrule
        \multirow{3}{*}{\parbox{2cm}{\centering Military\\Vehicles}}& 
    \multirow{1}{*}{Binary NN}&
    80.10\%&
    80.18\%\\
    &\multicolumn{1}{c}{DetRuleLearn~\cite{xi2023rulebased}} &
    83.45\%& 
    82.62\%\\
    &\multicolumn{1}{c}{f-EDR (ours)}&
    \textbf{84.08}\%& 
    \textbf{83.17}\%\\
    \cline{1-4} \multirow{3}{*}{ImageNet50}& 
    \multirow{1}{*}{Binary NN}& 
    72.85\%&
    68.96\%\\
    &\multicolumn{1}{c}{DetRuleLearn~\cite{xi2023rulebased}} &
    80.92\%& 
    72.78\%\\
    &\multicolumn{1}{c}{f-EDR (ours)} &
    \textbf{84.26}\%& 
    \textbf{77.78}\%\\
    \cline{1-4} \multirow{3}{*}{OpenImage36}& 
    \multirow{1}{*}{Binary NN}&
    64.80\%&
    63.65\%\\
    &\multicolumn{1}{c}{DetRuleLearn~\cite{xi2023rulebased}} &
    59.87\%& 
    46.46\%\\
    &\multicolumn{1}{c}{f-EDR (ours)} &
    \textbf{66.63}\%& 
    \textbf{65.83}\%\\
    \bottomrule
  \end{tabular}
  \label{tbl:error_detection_results}
\end{table}

\section{Experimental Results}
\label{sec:exp_results}
We experimented on three datasets in our work - our Military Vehicles dataset and subsets of $50$ classes from the ImageNet~\cite{russakovsky2015imagenet} dataset and $36$ classes from the OpenImage~\cite{Kuznetsova_2020} dataset. For each of those, we first trained a main model $\fthetahatmodel$ using the Binary Cross-Entropy (BCE) loss while employing a fitting State of the Art (SOTA) architecture and corresponding hyperparameters which we found to perform favorably among other architecture (the Vision Transformer~\cite{dosovitskiy2021an} b\_16 for the Military Vehicles and OpenImage and DINO V2~\cite{oquab2024dinov2} s\_14 for ImageNet). In the Military Vehicles dataset, there are a total of $9,444$ images and we took a ratio of $80:20$ between train and test. For the ImageNet50 dataset, we kept the original ratio from the paper of $1,300$ images in the train set and $50$ in the test set for each fine grain class, as well as $2000$ train and $400$ test images per fine grain class for the OpenImage36 dataset. In all datasets, we had ground truth constraints which we used to compute the fraction of samples violating ground truth constraints in the test set and/or determining the ability to recover constraints.

\noindent\textbf{Error Detection Results.} For error detection, we evaluated three approaches: Focused EDR (f-EDR, this paper), DetRuleLearn~\cite{xi2023rulebased}, and a neural (black-box) error prediction model inspired by related work (e.g., \cite{daftry_introspective_2016} - with an SOTA neural architecture) that effectively treats error detection as a binary classification problem.  For these neural error detection baselines, we used the same model architecture of our base model but retrained for this classification problem. Rules for both f-EDR and  DetRuleLearn were trained on the same training data used for the base models.  Conditions for the rules were derived from both the class of complementary granularity (e.g., fine for coarse predictions and vice-versa), lesser-performed models trained on the same data, and binary classification models for the same class (much in the same way additional models were used in \cite{xi2023rulebased}). We note that both f-EDR and DetRuleLearn used the same set of conditions to learn rules. Results of this experiment are shown in 
Table~\ref{tbl:error_detection_results} where we provide the balanced accuracy and F1-score of the total error - which is defined as applying a logical OR on all the per-class error classes. In all experiments, f-EDR significantly outperforms both DetRuleLearn and the neural-based error prediction by a significant margin in both metrics. These results corroborate the prescribed premise of the method in attempting to maximize for the performance gain of the error class.

\noindent\textbf{Constraint Recovery and Noise Tolerance.} We also experimented with scenarios involving noisy labels to demonstrate the noise tolerance of our method. In each experiment we pick a subset of fine-grain classes, remove their labels from the ground-truth of the training set and replace them with noisy labels from the remaining fine grain classes. We note that we omit these condition in the models used for conditions (and omit binary models for the removed classes altogether).  We experimented with levels of omitted fine grain classes and refer to this as the ``noise ratio'' (i.e., ratio of fine grain classes omitted to total fine grain classes available). We then train the whole pipeline from scratch, without any mention of the omitted labels. The results, depicted in Figure~\ref{fig:OOD} include an F1-score on the recovered constraints (compared to a ground truth set of constraints not otherwise used) as well as total error balanced accuracy and F1-score. Three key aspects of these result worth noting are: (1.) we can successfully recover the majority of HMC constraints with our approach, (2.) the recovery of constraints degrades gracefully with incomplete data, (3.) the prediction of errors also degrades gracefully with incomplete data.

\noindent\textbf{Improving Vision Model Performance.} A key envision use-case of this research is to use hierarchical constraints learned with f-EDR to improve the performance of a trained vision model. We compare our baseline model performance with a version whose training is constrained using f-EDR learned rules. Here we utilize rules learned that specify the relationships between fine and coarse grain classes as described in Section~\ref{sec:approach} and embed constraints in the training process using Logic Tensor Networks (LTN)~\cite{ltn22,LTNtorch} with an adjusted loss function~\cite{Vandenhende_2021} that considers both error loss and the level of consistency with f-EDR created constraints.  Results, shown in Table~\ref{tbl:error_correction_results}, indicate improved fine grain labels assignment in all cases; improved coarse-grain accuracy and reduced inconsistency (i.e., consistency of assigned classes with ground truth constraints) in all cases. These results show evidence that f-EDR can enhance model performance with a neurosymbolic approach - perhaps avoiding the pressing need for domain knowledge in certain situations.

\noindent\textbf{Conclusion.} In this paper, we presented a novel approach for detecting failures in HMC models that allows for the recovery of constraints along with error correction, in an explainable and scalable fashion. \noindent\textbf{Acknowledgement.} This research was Funded by ARO grant W911NF-24-1-0007.

\bibliographystyle{ACM-Reference-Format}
\balance
\bibliography{edcr-bib}

\end{document}